\pdfoutput=1

\documentclass[11pt]{article}

\usepackage[preprint]{acl}

\usepackage{times}
\usepackage{latexsym}

\usepackage[T1]{fontenc}

\usepackage[utf8]{inputenc}

\usepackage{microtype}

\usepackage{inconsolata}

\usepackage{graphicx}

\usepackage{booktabs}
\usepackage{xurl}

\title{DIY-MKG: An LLM-Based Polyglot Language Learning System}

\author{Kenan Tang \\
  UCSB \\
  \texttt{kenantang@ucsb.edu} \\\And
  Yanhong Li \\
  University of Chicago \\
  \texttt{yanhongli@uchicago.edu} \\\And
  Yao Qin \\
  UCSB \\
  \texttt{yaoqin@ucsb.edu} \\}

\begin{document}
\maketitle
\begin{abstract}
Existing language learning tools, even those powered by Large Language Models (LLMs), often lack support for polyglot learners to build linguistic connections across vocabularies in multiple languages, provide limited customization for individual learning paces or needs, and suffer from detrimental cognitive offloading.
To address these limitations, we design Do-It-Yourself Multilingual Knowledge Graph (DIY-MKG), an open-source system that supports polyglot language learning.
DIY-MKG allows the user to build personalized vocabulary knowledge graphs, which are constructed by selective expansion with related words suggested by an LLM. The system further enhances learning through rich annotation capabilities and an adaptive review module that leverages LLMs for dynamic, personalized quiz generation. In addition, DIY-MKG allows users to flag incorrect quiz questions, simultaneously increasing user engagement and providing a feedback loop for prompt refinement.
Our evaluation of LLM-based components in DIY-MKG shows that vocabulary expansion is reliable and fair across multiple languages, and that the generated quizzes are highly accurate, validating the robustness of DIY-MKG. \footnote{\url{https://github.com/kenantang/DIY-MKG}}
\end{abstract}

\section{Introduction}

\begin{figure}[htbp]
    \centering
    \includegraphics[width=1\linewidth]{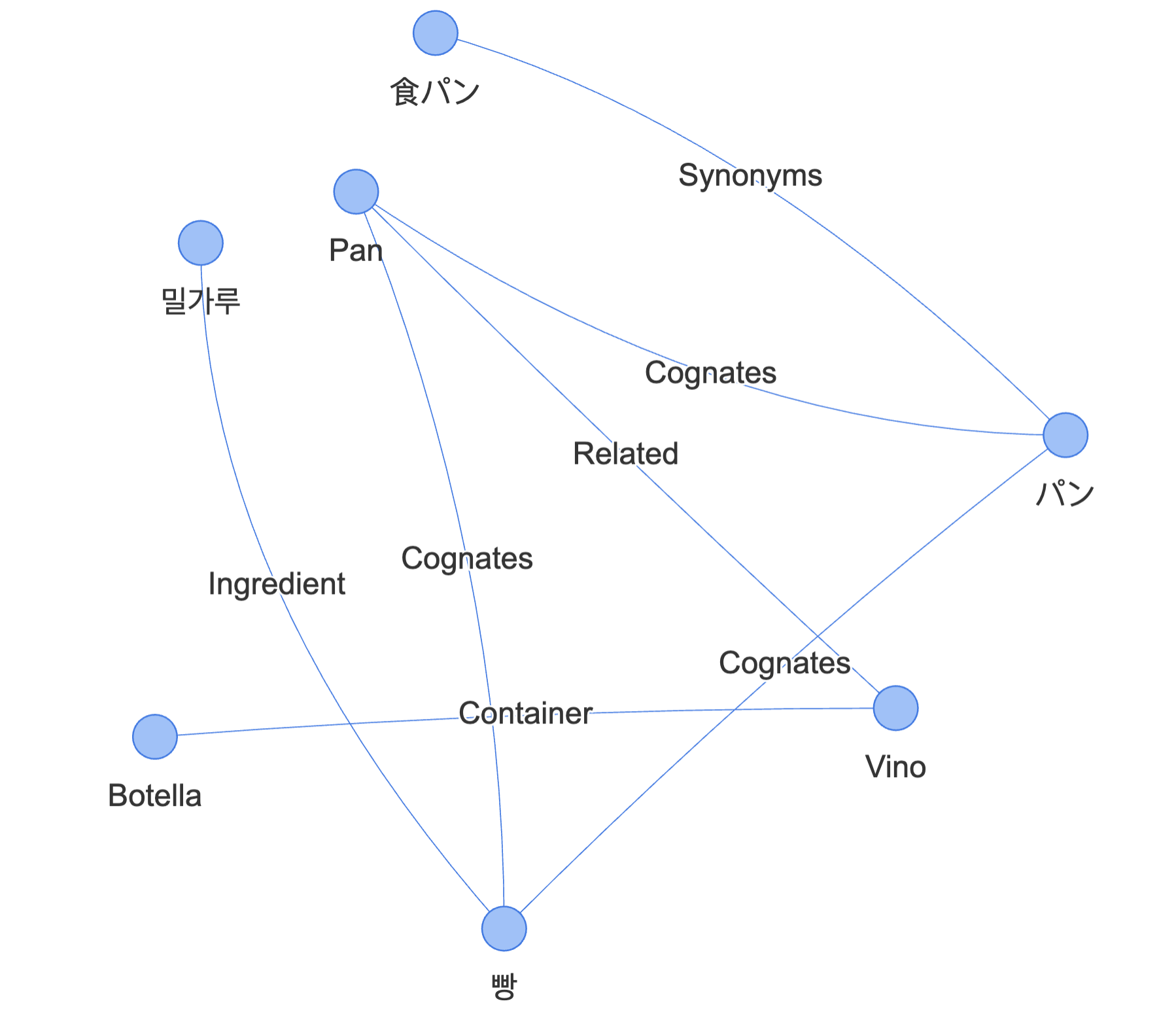}
    \caption{\textbf{DIY-MKG allows the language learner to construct a multilingual knowledge graph to help with vocabulary acquisition.} Linguistic connections between words, such as synonyms and cognates, enhance memorization of vocabulary knowledge. This is a screenshot taken from the DIY-MKG interface.}
    \label{fig:MKG}
\end{figure}

Large Language Models (LLMs) have demonstrated superior multilingual capabilities \cite{gemini_2.5, hurst2024gpt}. Therefore, LLMs have been used extensively in education applications, assisting language learners in improving capabilities such as reading, writing, and translation \cite{han2023recipe, han2023exploring, chu2025llm}.\footnote{In this paper, ``language learners'' refers specifically to learners who are learning a foreign language (L2, L3, etc.).}

However, less attention has been paid to using LLMs to assist vocabulary acquisition, a fundamental task for language learning \cite{barcroft2004second}. Currently available commercial software, such as Duolingo, usually includes a full-fledged system to help vocabulary acquisition. Treating the user's current vocabulary as a word list, a commercial software tracks the learning progress, reminds the user to review words, and provides interactive lessons and quizzes to assist vocabulary acquisition.

Despite the maturity of commercial software, LLMs open up exciting new possibilities for vocabulary acquisition, which have not yet been incorporated into popular commercial software. With the strong capabilities of LLMs in mind, we identify three limitations of existing systems. First, existing systems do not exploit linguistic connections across words in multiple languages (examples in Figure \ref{fig:MKG}). Secondly, existing systems are not customized to the pace of diverse language learners (Section \ref{sec:diy-mkg-adaptive-reviewing}), as the systems heavily rely on predefined lessons and quizzes. Thirdly, when LLMs are involved in language learning, many systems lack a well-structured design that balances the contributions of both the language learner and the LLM to the learning process. When the learner blindly accepts the material generated by an LLM, the learning process is hampered by excessive \textit{cognitive offloading} \cite{kosmyna2025your}, where the learner fails to engage in independent problem-solving or critical thinking while interacting with the LLM.

To address these three limitations, we propose \textbf{Do-It-Yourself Multilingual Knowledge Graph} (DIY-MKG), an open-source and customizable system to support language learners. DIY-MKG is an interface that allows a language learner to save vocabulary knowledge in the form of a knowledge graph. In the interface, the language learner can expand their vocabulary, add personalized annotation to the knowledge graph, and test their own knowledge with automatically generated quizzes, all with the structured assistance of an LLM.

DIY-MKG follows three key design principles. First, DIY-MKG supports multiple languages, with an emphasis on drawing linguistic connections between words in multiple languages (Figure \ref{fig:MKG}), which heavily contributes to vocabulary acquisition (Section \ref{sec:related-work}). Second, DIY-MKG is open-source, allowing customization of all of its components such that the language learner can adapt the system to specific domains or learning stages. Third, DIY-MKG discourages cognitive offloading by providing a checkbox-based interface, enabling users to easily label problematic responses generated by the LLM. Taken together, these design principles strategically enhance the learning experience of language learners who speak multiple languages.

Our contributions can be summarized as follows:
\begin{itemize}
    \item We release DIY-MKG, an MIT-licensed system for polyglot language learning, with a novel emphasis on vocabulary acquisition.
    \item We evaluate LLM-based components in DIY-MKG, ensuring its reliability.
\end{itemize}

In the following sections, we first introduce the related work that motivate DIY-MKG. Then, we explain the main functionalities of DIY-MKG. Next, we evaluate LLM-based components in DIY-MKG. Finally, we conclude and discuss future directions. 

\section{Related Work}
\label{sec:related-work}

Vocabulary acquisition is a fundamental part of second language acquisition \cite{barcroft2004second}. Strong vocabulary knowledge contributes to many aspects of language proficiency \cite{hsueh2000unknown, sun2023affects}. Thus, extending the vocabulary size, specifically vocabulary depth and breadth \cite{qian1999assessing, qian2002investigating, schmitt2014size}, is of high priority for language learners.

Furthermore, for polyglots, vocabulary knowledge from the known languages helps vocabulary acquisition in a new language \cite{bartolotti2017bilinguals}. Vocabulary knowledge is shared across languages in the form of one-to-one cognates \cite{garcia2025cognate, nagy1993spanish, sanahuja2024impact, xiong2020time} or multiple words that share common roots, prefixes, or suffixes \cite{jeon2011contribution, zhang2024unpacking, crosson2016middle,crosson2019extending}. The sharing of vocabulary knowledge exists universally in language pairs and even triplets \cite{choi-etal-2004-korean, shen2022modern, heinrich2020language}, making this vocabulary acquisition strategy available for a wide range of language learners. 

Despite ample evidence of multilingual vocabulary knowledge helping language learning, previous work on LLM-assisted language learning mostly focuses only on higher-level tasks such as reading or writing \cite{han2023recipe, han2023exploring, chu2025llm}, without considering the lower-level task of vocabulary acquisition. Also, previous work mainly consider second language acquisition (SLA), where LLMs are trained or tested for students speaking one language and learning a second language, e.g., native Korean speakers who are learning English \cite{han-etal-2024-recipe4u, han-etal-2024-llm}. Even popular commercial software, such as Duolingo, is commonly designed only around SLA. With existing systems, polyglots seldom benefit from their extra knowledge when learning a new language.

Hence, we fill the gap by proposing DIY-MKG, an LLM-driven system inspired by various vocabulary acquisition strategies \cite{brown1991comparison, ellis1995psychology, lawson1996vocabulary}. With DIY-MKG, a language learner can fully exploit their prior knowledge in multiple languages towards learning a new language, with customizable and reliable assistance from LLMs.

\begin{figure*}[htbp]
    \centering
    \frame{\includegraphics[width=1\linewidth]{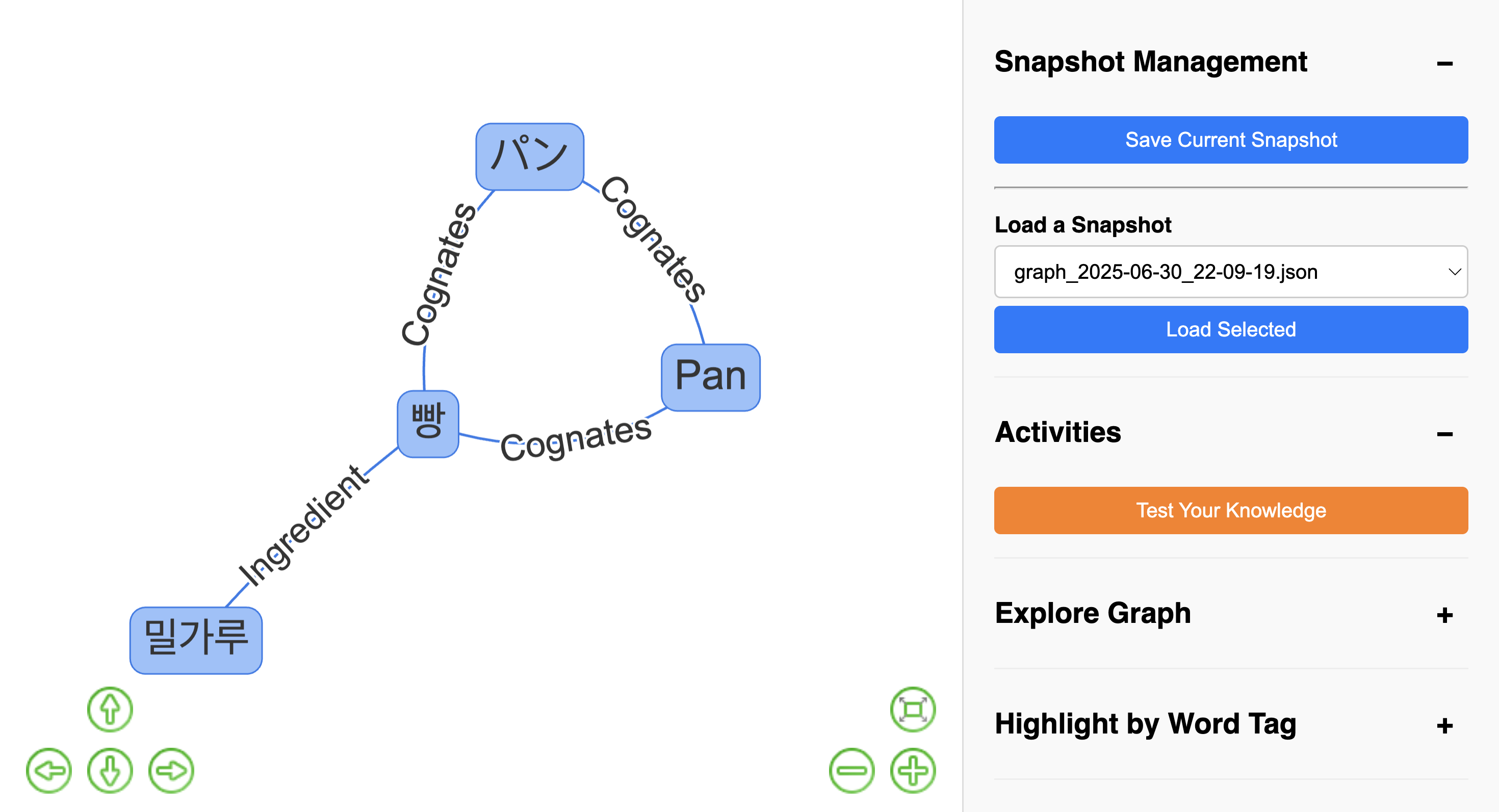}}
    \caption{\textbf{A screenshot of DIY-MKG, taken after zooming into a chosen word.} The subgraph of words connected to the chosen word is visualized on the left. On the right, the side panel supports multiple functionalities (Section \ref{sec:diy-mkg}).}
    \label{fig:diy-mkg}
\end{figure*}

\section{DIY-MKG}
\label{sec:diy-mkg}

In this section, we introduce functionalities of DIY-MKG (Figure \ref{fig:diy-mkg}). The functionalities are organized into 3 main categories, namely vocabulary expansion, rich annotations, and adaptive reviewing.

\subsection{Vocabulary Expansion}
\label{sec:diy-mkg-vocabulary-expansion}

DIY-MKG assists the user to gradually expand their vocabulary during their learning process. Different from other software that treats vocabulary as a list, DIY-MKG saves and visualizes the vocabulary as a multilingual knowledge graph, a data structure that better supports vocabulary acquisition strategies based on linguistic connection between words. 

\begin{figure}[htbp]
    \centering
    \frame{\includegraphics[width=0.7\linewidth]{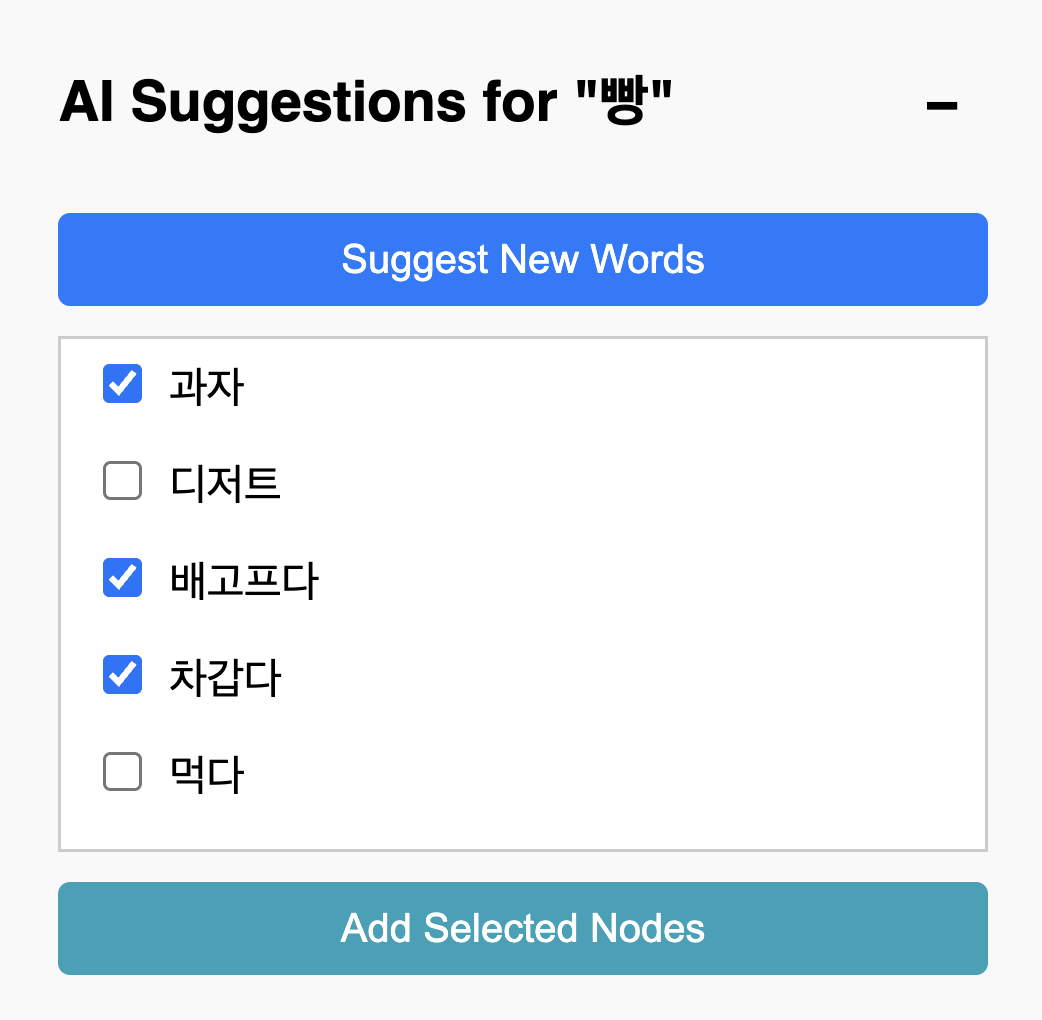}}
    \caption{\textbf{DIY-MKG allows the user to expand their vocabulary by selecting related words recommended by an LLM.} The selection step prevents the user from blindly accepting recommendations of the LLM, mitigating detrimental effects of cognitive offloading.}
    \label{fig:suggestions}
\end{figure}

\paragraph{Knowledge Graph Construction} To start constructing their own knowledge graph, the user can manually add a set of words they already know into the vocabulary. The words will be used as the initial nodes in the knowledge graph. Then, in the interface, the user can zoom into a certain word by clicking or typing the word. Based on this chosen word, the user can query an LLM to generate related words, which are going to be added selectively to the vocabulary (Figure \ref{fig:suggestions}). Thanks to the strong prompt following ability of LLMs, the related words can include synonyms, antonyms, words with similar spelling, words of a similar difficulty level, etc. More importantly, related words can also be retrieved from other languages by using different prompts. This functionality distinguishes DIY-MKG from traditional dictionaries or knowledge graphs \cite{miller1995wordnet}, where connections between words are commonly monolingual, predefined, and static.

\paragraph{Selective Expansion} After the related words are generated by LLMs, DIY-MKG requires the user to manually select from the related words. The selected words will be then automatically connected to the chosen word and added into the vocabulary. This design prevents the user from fully relying on LLMs without critical thinking. By manually selecting words, the user maintains full control over the vocabulary expansion, while benefiting from the high creativity and deep vocabulary knowledge of LLMs \cite{tang-etal-2024-creative}. 

\paragraph{Safety Guardrails} LLMs sometimes generate inappropriate content for certain user groups, such as young children. To improve the safety of DIY-MKG, we design the following two guardrails. First, DIY-MKG supports both API-based LLMs and local LLMs. Compared to API-based LLMs, local LLMs with stronger safety guarantees could be chosen for specific user groups. Secondly, DIY-MKG supports a safe mode, where the generated list of related words will be filtered by an additional LLM query to ensure appropriateness (prompt in Appendix \ref{sec:prompts}). These two safety guardrails can flexibly support diverse safety requirements for different user groups.

\paragraph{High Customizability} Since DIY-MKG is lightweight and fully open-source, functionalities above, including vocabulary expansion prompts and safety guardrails, can be easily customized to address user needs. For example, if a user would like to learn vocabulary in the medical field, they can (1) use a vocabulary expansion prompt that asks for domain-specific words, (2) use a fine-tuned LLM with better medical knowledge than general purpose LLMs, and (3) choose a filtering prompt that is not oversensitive to medical terms. A user can also update these components at different stages of language learning to accommodate for their improved vocabulary knowledge.

\subsection{Rich Annotations}

To conveniently update and review multilingual vocabulary knowledge, a user requires more than a vanilla knowledge graph with only node and edge labels. Hence, we design DIY-MKG to support rich annotations at the following three levels.

\paragraph{Node Level} At the node level, the user can save any information related to a single word. Examples include definition, example sentences, or specific context where the word is encountered. The information can be edited at the side panel and can be visualized by hovering over the word. Markdown format is supported for node-level annotation. Moreover, the user can add custom tags to each word. The custom tags can be activated at the side panel so that all tagged words in the whole knowledge graph are highlighted. 

\begin{figure}
    \centering
    \frame{\includegraphics[width=1\linewidth]{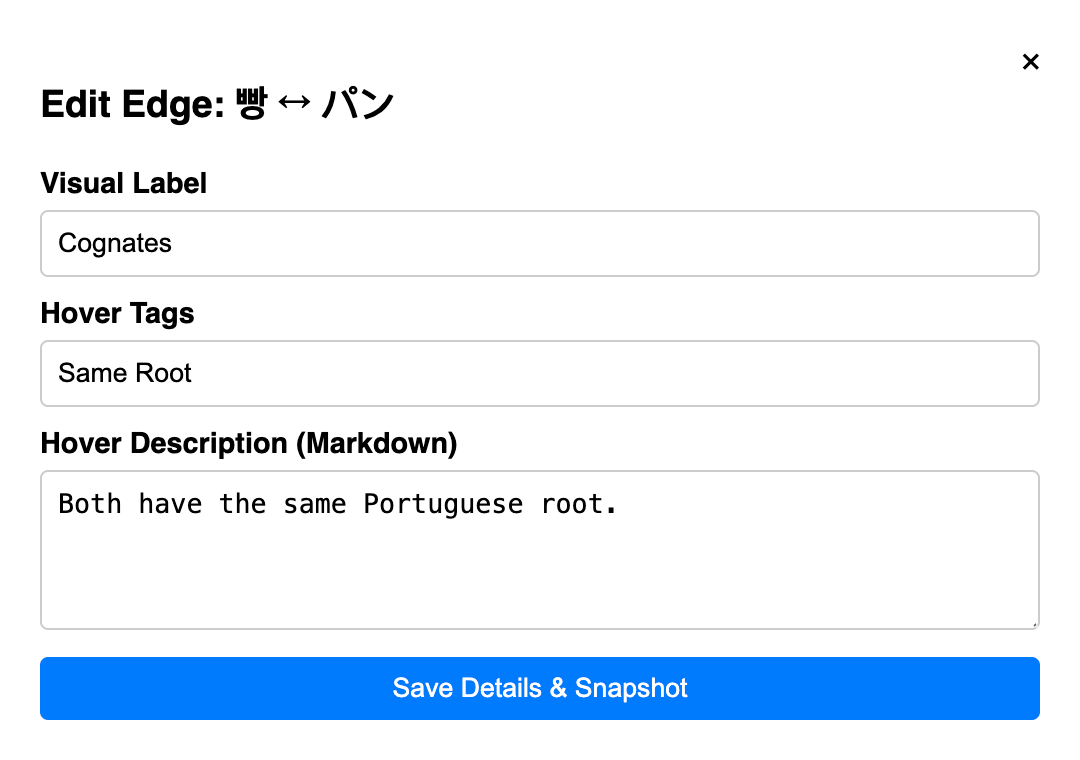}}
    \caption{\textbf{DIY-MKG allows the user to provide rich annotations at the edge level.} The user can provide a visual label, multiple hover tags, or a hover description of custom length. Markdown rendering is supported in the interface, enabling elaborate edge annotations.}
    \label{fig:edge-editing}
\end{figure}

\paragraph{Edge Level} At the edge level, the user can similarly save any information related to a pair of words (Figure \ref{fig:edge-editing}). Examples include explanations for cognates, for words with similar roots, or for any personalized connection that could be drawn between two words. Markdown format, hovering preview, and edge tags are also supported for edge-level annotation. While the nodes and edges can be added via vocabulary expansion (Section \ref{sec:diy-mkg-vocabulary-expansion}), a user can always freely add or remove nodes and edges using the side panel. 

\paragraph{Hyper-Edge Level} At the hyper-edge level, the user can link multiple words by a document. Examples include a story that is written using words of a specific difficulty level, a blog post that explains cognates in different languages, or a quiz that tests vocabulary knowledge. Currently, DIY-MKG supports saving hyper-edge information from a quiz (Section \ref{sec:diy-mkg-adaptive-reviewing}) as local documents. A document-type-specific support for hyper-edge visualization will be added in the future (Appendix \ref{sec:hyper-edge-visualization}).

All three levels of annotations are saved locally in human-readable JSON files. A user can freely edit any part of the knowledge graph outside the interface, or they can export the knowledge graph for additional analyses. DIY-MKG also supports saving and loading snapshots of the knowledge graph, enabling convenient version control.

\subsection{Adaptive Reviewing}
\label{sec:diy-mkg-adaptive-reviewing}

After constructing a knowledge graph of words, a user also needs to frequently review the vocabulary to enhance memorization. The key to effective reviewing is adapting the reviewing process to the less‑frequently used words in the vocabulary. In DIY-MKG, we design the following functionalities to provide a adaptive reviewing experience.

\paragraph{Click Counter} In DIY-MKG, the number of times that a user clicks a word (node) is saved as an attribute of the word. When a word is clicked, the user is either updating the node-level annotation for the word or expanding the vocabulary based on the word. Hence, the click count of each word serves as a good proxy for the user's level of understanding and memorization of the word. The words with the lowest click counts are the ones that need more frequent review.\footnote{For certain users, words with the lowest click counts could be ones that are easy to memorize, requiring less frequent review. Thus, DIY-MKG provides click counts as a statistic to help users customize their reviewing needs.}

\paragraph{Quiz Generation} DIY-MKG supports a quiz generation functionality for reviewing vocabulary knowledge. After the user clicks the ``Test My Knowledge'' button (Figure \ref{fig:diy-mkg}), a quiz with multiple-choice questions and fill-in-the-blank questions will be generated based on the lowest-frequency words. The quiz is automatically generated with an LLM, so fresh questions can be generated to prevent the user from memorizing shortcuts. A quiz example can be found in Appendix \ref{sec:quiz-example}. After the user completes the quiz, the quiz will be automatically graded by matching the user's answer string with the correct answer string. Finally, the quiz results will be saved locally for future reference. 

\paragraph{Question Flagging} Since LLM-generated question-answer pairs can sometimes be incorrect, DIY-MKG further allows the user to flag incorrect question-answer pairs after they submit the quiz answer and see the results. The flagged questions are labeled in the local quiz file, which can be used to iteratively improve the quiz generation prompt in future versions of DIY-MKG. The question flagging functionality is also designed to improve user engagement with the quiz and mitigating cognitive offloading \cite{kosmyna2025your}.

\paragraph{High Adaptability} We would like to highlight the higher adaptability of the reviewing process in DIY-MKG, compared to that in the other commercial software. While commercial software often support more diverse reviewing methods, the user cannot specify the set of words that need to be reviewed or how frequently each word should be reviewed. These two shortcomings often lead to an undesirable reviewing schedule, in which words that are already well-memorized are presented too frequently, whereas some new words are never reviewed at all and are thus forgotten by the user. In contrast, in DIY-MKG, a user can implement their own review schedule based on node-level statistics, adapting the schedule to personalized learning curves. This feature is particularly useful when the vocabulary consists of domain-specific words, the forgetting curve of which differs from that of common words \cite{zaidi2020adaptive}. 

\section{Evaluation}
\label{sec:evaluation}

To examine if the proposed functionalities can work as intended, we evaluate vocabulary expansion and adaptive reviewing, the two LLM-based components in DIY-MKG. All experiments are conducted using \texttt{Llama-3.3-70B-Instruct} with a temperature of 0. We release the evaluation script and our evaluation data to allow reproduction of our results and evaluation on other models.

\subsection{Vocabulary Expansion}
\label{sec:evaluation-vocabulary-expansion}

To evaluate vocabulary expansion, we test if the provided prompt (Appendix \ref{sec:prompts}) can help iteratively expand the vocabulary with new words. We evaluate under a monolingual setting (i.e., no related words from another language) and take the following steps. First, we specify a random word in the language as the starting word. Then, we apply the prompt to generate words related to the specified word in the same language. Among the generated words, if a word is not yet in the vocabulary, we add the word into the vocabulary, together with an LLM-generated description (prompt in Appendix \ref{sec:prompts}). Next, we sample a word from the vocabulary which has not been previously used for expansion. This word will be used to repeat the process above. 

We evaluate the prompt for Spanish, Korean, and Japanese words. For each language, we randomly sample 10 words as the starting word. For each starting word, we iteratively apply the prompt 500 times. Then, we examine how the vocabulary size increases as the number of iterations increase. 

\begin{figure}[htbp]
    \centering
    \includegraphics[width=\linewidth]{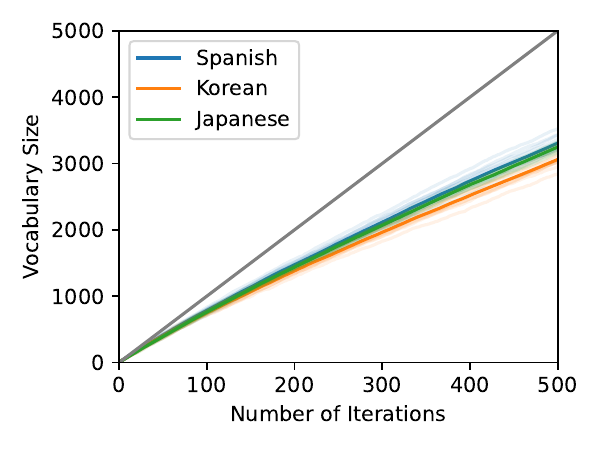}
    \caption{\textbf{DIY-MKG reliably expands the vocabulary with low sensitivity to the language and the starting word.}  For each language, 10 opaque curves represent vocabulary sizes obtained from 10 different starting words, and one solid curve represents the average of the 10 curves. The standard deviations across languages and starting words are small, indicating fairness for all language learners. The diagonal gray line represents the upper-bound of vocabulary size, assuming all words in each iteration have not yet appeared in the vocabulary. While the real expansion rate is lower, the vocabulary does not saturate after 500 iterations. Therefore, DIY-MKG can reliably help the language learner to expand their vocabulary by consistently introducing new words.}
    \label{fig:vocabulary-expansion}
\end{figure}

\paragraph{High Reliability} Figure \ref{fig:vocabulary-expansion} shows that the prompt reliably expands the vocabulary. The diagonal gray line represents the upper-bound of vocabulary size, assuming all words in each iteration have not yet appeared in the vocabulary. While the real expansion rate is lower due to duplicate words, the vocabulary does not saturate after 500 iterations. The final vocabulary sizes are around 3,000, which is comparable to the total vocabulary size (2,114) of the English version of the Korean course on Duolingo. Hence, a language learner can use DIY-MKG for a long time and still consistently learn new words.

\paragraph{Fairness} A low sensitivity to the language and starting word ensures the fairness of DIY-MKG for all language learners. On the one hand, the standard deviation of the average vocabulary sizes across languages are small, indicating fairness for learners of different languages. On the other hand, the standard deviations of the vocabulary sizes across different starting words in the same language are small. This suggests that the learning experience will be similar regardless of the starting word chosen by the user, further ensuring fairness. 

\subsection{Adaptive Reviewing}
\label{sec:evaluation-adaptive-reviewing}

To evaluate adaptive reviewing, we conduct human study on the generated multiple-choice questions and fill-in-the-blank questions. We generate 50 multiple-choice questions and 50 fill-in-the-blank questions in Spanish, Korean, and Japanese, totaling 300 questions. The prompt for generating the questions are the same as ones used in the system (i.e., the questions will be generated in a JSON format), which can be found in Appendix \ref{sec:prompts}. The words for generating the questions are randomly selected from the vocabulary generated from the first part of the evaluation. The correctness of the question-answer pairs are judged by \texttt{gpt-4.1-2025-04-14} with a temperature of 0. The judge prompt can be found in Appendix \ref{sec:prompts}.

\paragraph{Variable Correctness} Table \ref{tab:adaptive-reviewing} shows the variable correctness of the generated questions. While multiple-choice question-answer pairs are almost always correct, the fill-in-the-blank question-answer pairs show fluctuating correctness for the three tested languages. Upon manual inspection, the incorrect fill-in-the-blank question-answer pairs usually have a question that is ambiguous and thus cannot be answered (Appendix \ref{sec:quiz-example}). Hence, we design the question flagging functionality (Section \ref{sec:diy-mkg-adaptive-reviewing}) to handle these ambiguous questions.

\begin{table}[htbp]
    \centering
    \begin{tabular}{ccc}
        \toprule
        Language & MCQ & FIB \\
        \midrule
        Spanish & 98\% & 82\% \\
        Korean & 98\% & 84\% \\
        Japanese & 98\% & 76\% \\
        \bottomrule
    \end{tabular}
    \caption{\textbf{The correctness of generated question-answer pairs varies with language and question type.} While multiple-choice questions (MCQ) are almost always correct in all three languages, fill-in-the-blank questions (FIB) show fluctuating correctness for the three tested languages. Hence, we incorporate an error flagging mechanism in DIY-MKG to facilitate the search for failure cases and for better prompts. }
    \label{tab:adaptive-reviewing}
\end{table}

\section{Conclusion and Future Work}

In this paper, we introduce DIY-MKG, a support system for polyglot language learning. DIY-MKG is carefully designed to ensure a customizable multilingual experience, where the language learner interacts with an LLM in a structured and thoughtful way. In the future, we will extend the interface to more modalities, including audio and images. Moreover, while DIY-MKG is now designed only for language learning, we plan to adapt the framework to other disciplines, where knowledge graphs can be similarly used to encode the connection between concepts. Finally, the learning traces obtained from DIY-MKG will shed light on how language learners adjust their learning process in the presence of LLMs, facilitating future research in education and human-LLM interaction.

\section*{Limitations}

DIY-MKG has the following limitations:

\begin{enumerate}
    \item The current UI is a research preview. We will keep updating the UI according to user feedback. However, the layout of the UI is designed for a computer screen. Since the information density in the interface is designed to be high, we will likely not support a mobile version of DIY-MKG.
    \item To ensure best experience, a language learner needs to use a powerful multilingual language model for DIY-MKG. This could be costly for some language learners. Furthermore, for low-resource languages, there might not yet exist a satisfactory model. 
    \item We have not yet conducted a large-scale user study. While knowledge graph visualization and editing will always remain as core functionalities of DIY-MKG, the other components are subject to future changes.
\end{enumerate}

We are actively developing new versions of DIY-MKG to address the above limitations.

\bibliography{anthology,custom}

\appendix

\section{Prompts}
\label{sec:prompts}

\begin{figure}[htbp]
    \centering
    \includegraphics[width=1\linewidth]{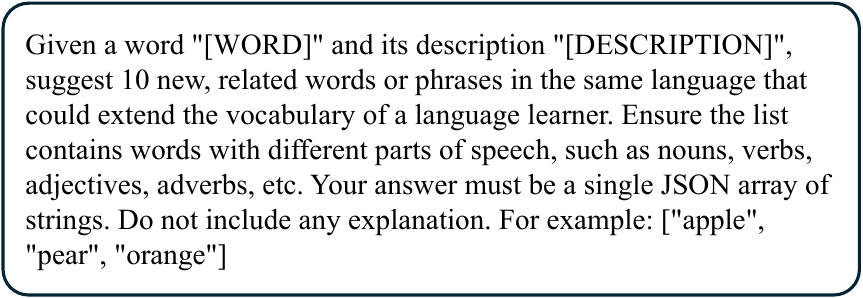}
    \caption{Prompt for suggesting related words.}
    \label{fig:prompt-suggest}
\end{figure}

\begin{figure}[htbp]
    \centering
    \includegraphics[width=1\linewidth]{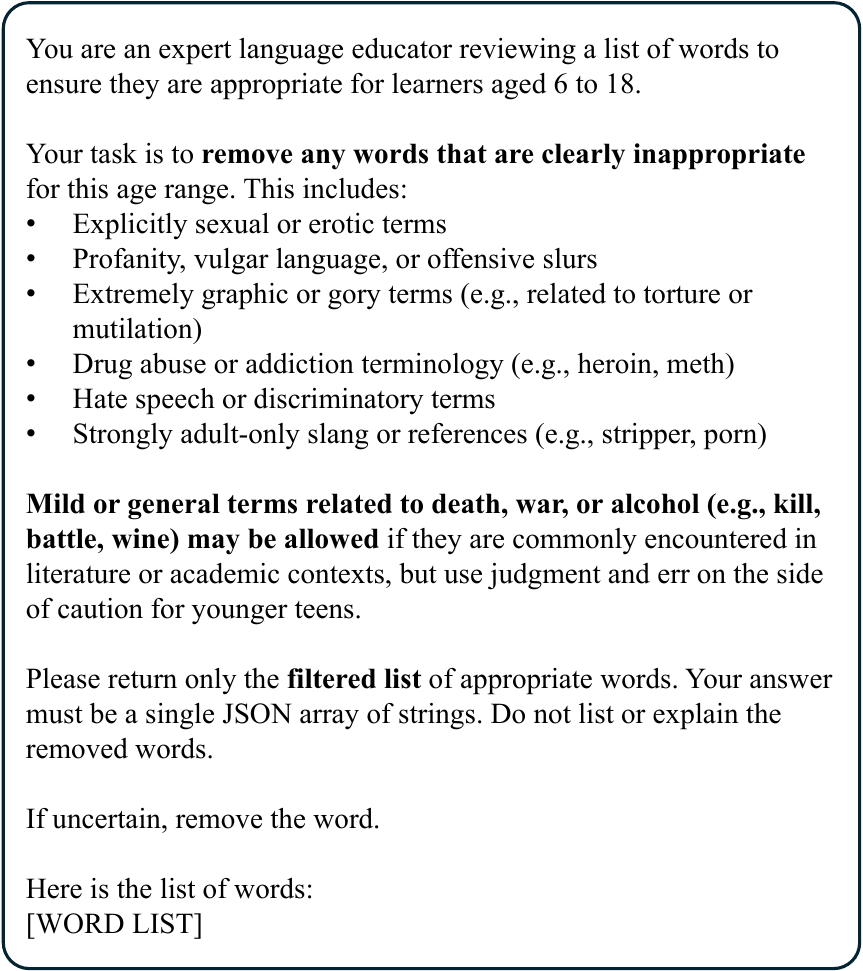}
    \caption{Prompt for filtering inappropriate words.}
    \label{fig:prompt-filter}
\end{figure}

\begin{figure}[htbp]
    \centering
    \includegraphics[width=1\linewidth]{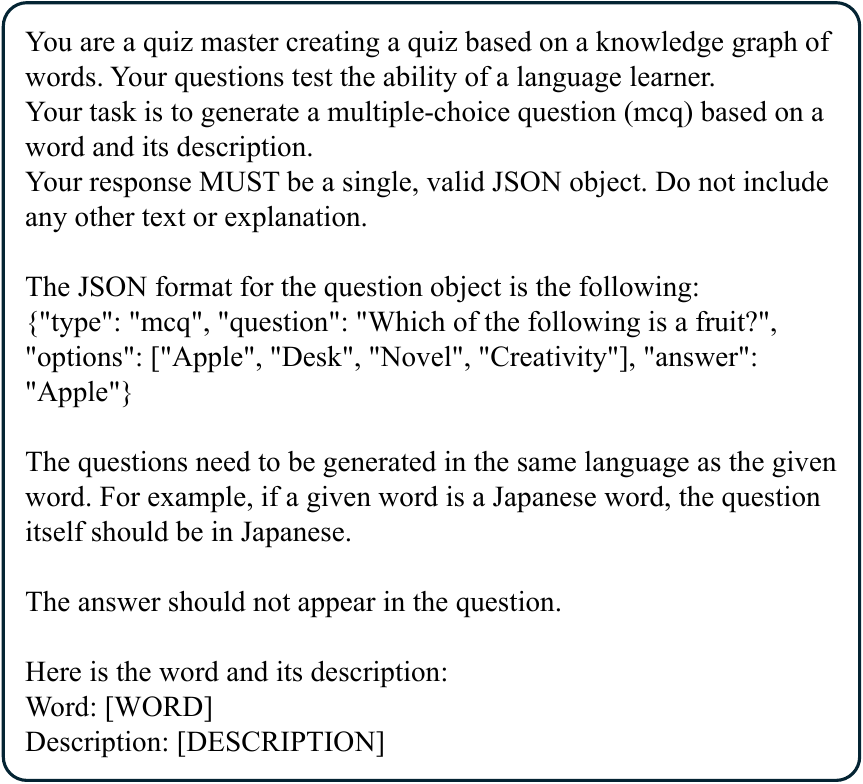}
    \caption{Prompt for generating multiple-choice questions.}
    \label{fig:prompt-mcq}
\end{figure}

\begin{figure}[htbp]
    \centering
    \includegraphics[width=1\linewidth]{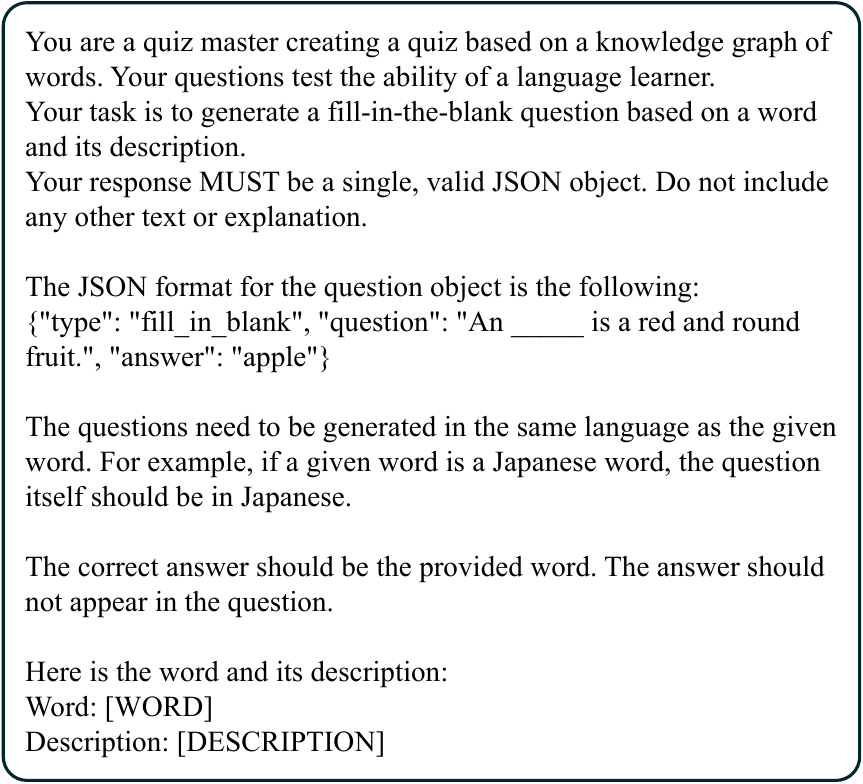}
    \caption{Prompt for generating fill-in-the-blank questions.}
    \label{fig:prompt-fib}
\end{figure}

In this section, we list the prompts for suggesting related words (Figure \ref{fig:prompt-suggest}), filtering inappropriate words (Figure \ref{fig:prompt-filter}), generating multiple-choice questions (Figure \ref{fig:prompt-mcq}), and generating fill-in-the-blank questions (Figure \ref{fig:prompt-fib}). These prompts are used in the interface by default, and the user can also customize these prompts for specific needs.

\begin{figure}[htbp]
    \centering
    \includegraphics[width=1\linewidth]{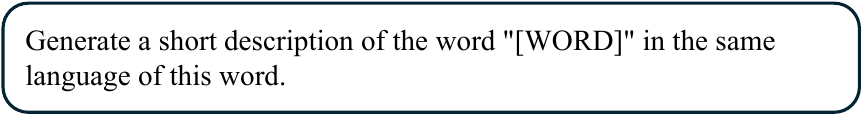}
    \caption{Prompt for generating descriptions.}
    \label{fig:prompt-describe}
\end{figure}

\begin{figure}[htbp]
    \centering
    \includegraphics[width=1\linewidth]{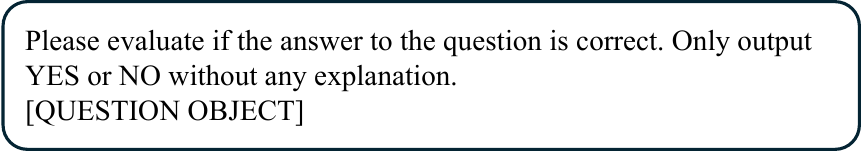}
    \caption{Prompt for evaluating question-answer pairs.}
    \label{fig:prompt-evaluate}
\end{figure}

Additionally, we list the prompts for generating descriptions (Figure \ref{fig:prompt-describe}) and evaluating question-answer pairs (Figure \ref{fig:prompt-evaluate}), which are used in Section \ref{sec:evaluation} for evaluation. These prompts are not used in the interface.

\section{Hyper-Edge Visualization}
\label{sec:hyper-edge-visualization}

\begin{figure}
    \centering
    \includegraphics[width=1\linewidth]{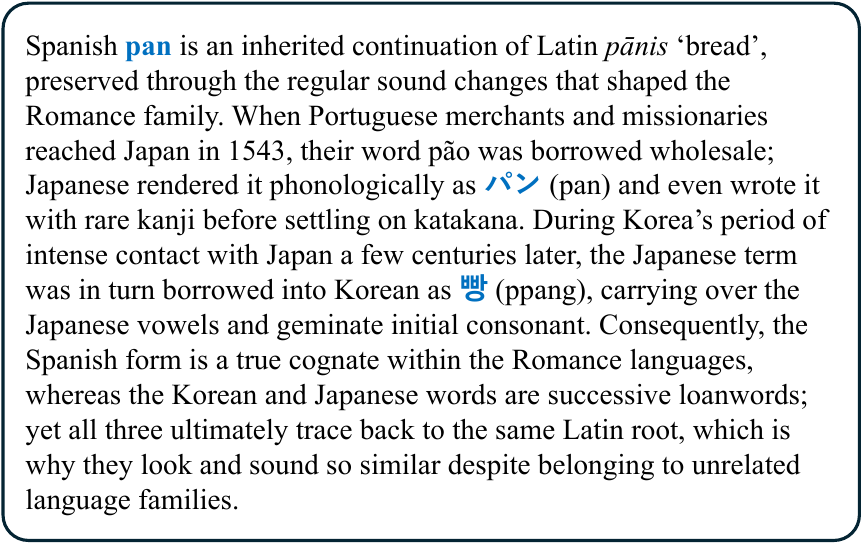}
    \caption{\textbf{A visualization of a hyper-edge that connects words in Spanish, Korean, and Japanese.} The connected words are highlighted in blue.}
    \label{fig:hyper-edge}
\end{figure}

Hyper-edges in a graph can be visualized in diverse ways.\footnote{Please see \url{https://github.com/iMoonLab/DeepHypergraph} or \url{https://xgi.readthedocs.io/en/stable/api/tutorials/focus_5.html} for examples of hyper-edge visualization.} However, to adapt to the need of language learners, we do not consider direct visualization on the graph, but instead visualize hyper-edges as a single document, where words (nodes) connected by the hyper-edge are highlighted in the document. An example is shown in Figure \ref{fig:hyper-edge}. The document is generated by \texttt{o3} in the ChatGPT interface. Hyper-edge visualization will be implemented in DIY-MKG in the future.

\section{Quiz Example}
\label{sec:quiz-example}

\begin{figure*}
    \centering
    \includegraphics[width=0.7\linewidth]{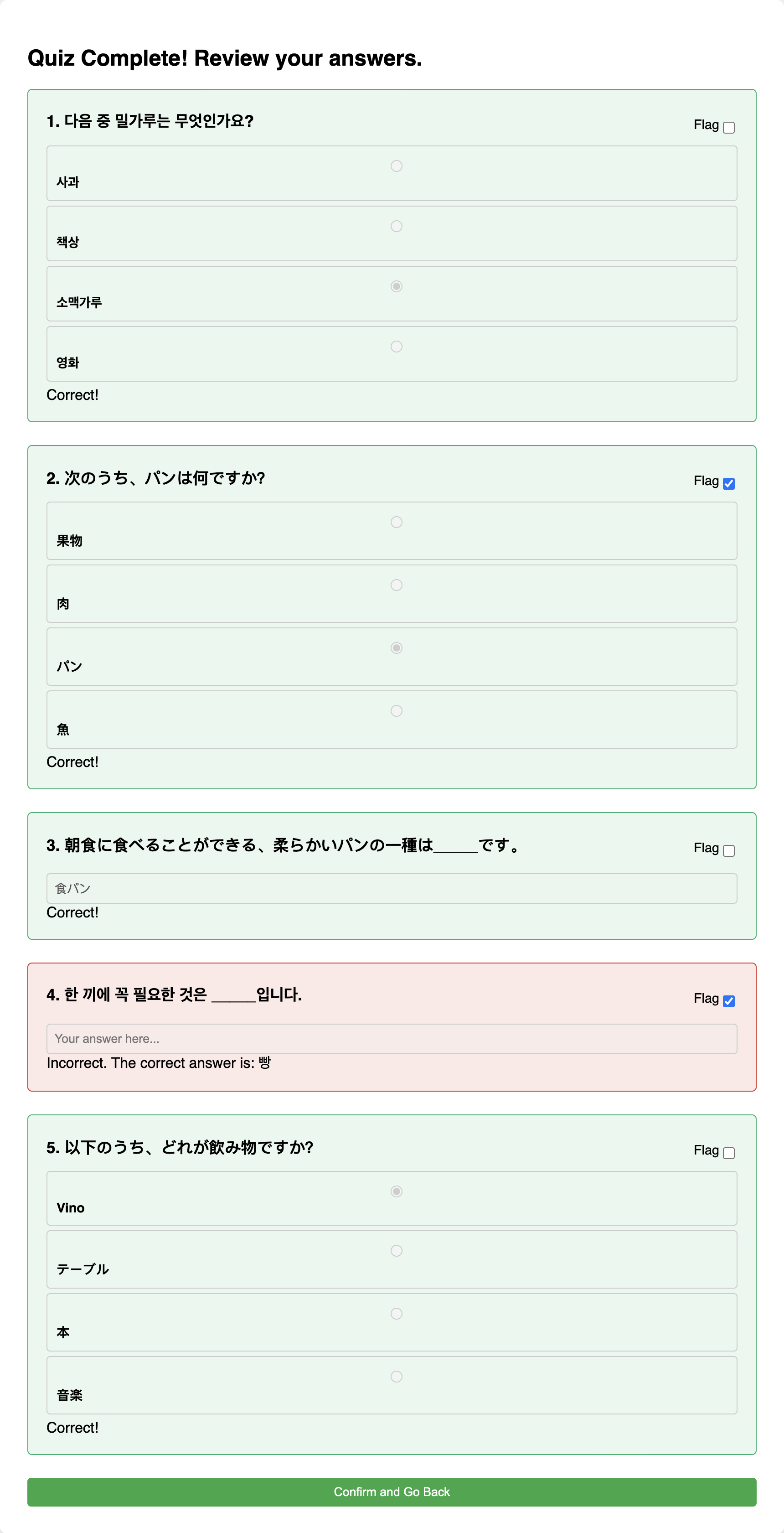}
    \caption{A quiz example consisting of 2 multiple-choice questions and 3 fill-in-the-blank questions.}
    \label{fig:quiz-example}
\end{figure*}

A quiz example is shown in Figure \ref{fig:quiz-example}. The quiz consists of 2 multiple-choice questions and 3 fill-in-the-blank questions, automatically generated by \texttt{gpt-4o-mini-2024-07-18}. After the user submits the answer, DIY-MKG checks the correctness of the answer, highlighting correct user responses in green and incorrect user responses in red. However, the second question is a tautology, and the fourth question is ambiguous. In this case, the user can flag the questions as incorrect, since they do not meaningfully test vocabulary knowledge. The questions, correct answers, user responses, and user flags will all be saved locally in a JSON format after clicking the ``Confirm and Go Back'' button.

\end{document}